%% file: main.tex
\documentclass[runningheads]{llncs}

\usepackage{eccv}

\usepackage{eccvabbrv}

\usepackage{graphicx}
\usepackage{booktabs}
\usepackage{array}
\usepackage{twemojis}

\usepackage[accsupp]{axessibility}  

\usepackage[pagebackref,breaklinks,colorlinks]{hyperref}

\usepackage{orcidlink}

\begin{document}

\title{CamFreeDiff: Camera-free Image to Panorama Generation with Diffusion Model} 


\author{Xiaoding Yuan\inst{1,2} \and
Shitao Tang\inst{3} \and
Kejie Li\inst{2} \and Alan Yuille\inst{1} \and Peng Wang\inst{2}}


\institute{The Johns Hopkins University \and
ByteDance
\and
Simon Fraser University}

\maketitle

\input{text/0-abstract}
\input{text/1-introduction}
\input{text/2-related_work}

\input{text/3-preliminary}

\input{text/4-method}
\input{text/5-experiments}

\input{text/6-conclusion}

%
%
\bibliographystyle{splncs04}
\bibliography{egbib}
\end{document}

%% file: text/0-abstract.tex
\begin{figure}[]
  \centering
  \includegraphics[width=\textwidth]{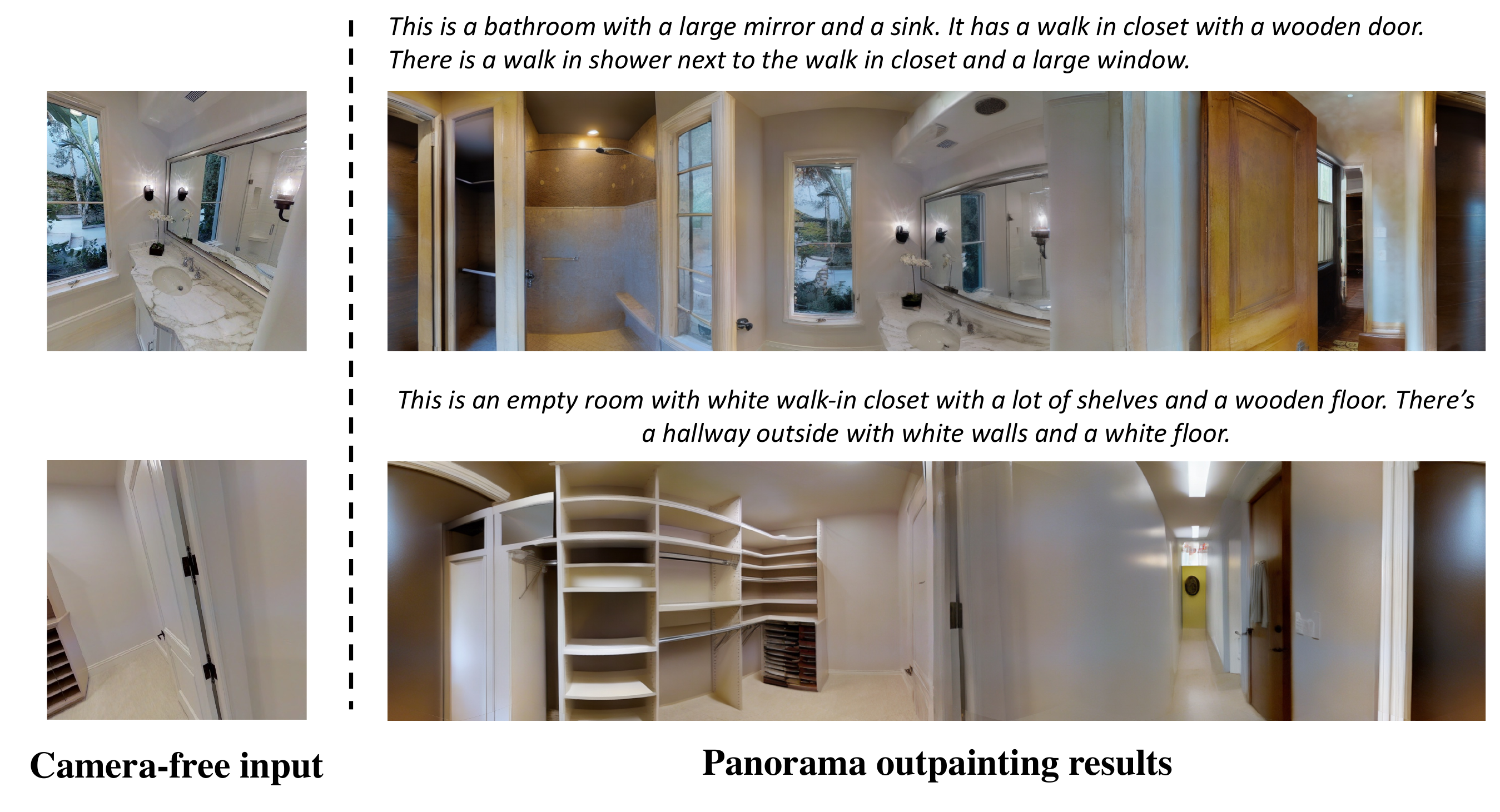} 
  \caption{Camera-free image-to-panorama generation: our method generates panoramas with reasonable layout and canonical viewpoint with unknown input image camera parameters.}
  \label{fig:1}
\end{figure}

\begin{abstract}
  This paper introduces Camera-free Diffusion (CamFreeDiff) model for 360-degree image outpainting from a single camera-free image and text description. This method distinguishes itself from existing strategies, such as MVDiffusion, by eliminating the requirement for predefined camera poses. Instead, our model incorporates a mechanism for predicting homography directly within the multi-view diffusion framework. The core of our approach is to formulate camera estimation by predicting the homography transformation from the input view to a predefined canonical view. The homography provides point-level correspondences between the input image and targeting panoramic images, allowing connections enforced by correspondence-aware attention in a fully differentiable manner. Qualitative and quantitative experimental results demonstrate our model's strong robustness and generalization ability for 360-degree image outpainting in the challenging context of camera-free inputs. 
\end{abstract}

%% file: text/1-introduction.tex

\section{Introduction}
Unlike traditional movies, 360-degree movies create an immersive experience that allows viewers to feel as if they are part of the movies's environment rather than merely observing it from a fixed perspective. This immersive aspect has been significantly enhanced with the advent and proliferation of AR/VR devices like VisionPro. Creating 360-degree movies typically requires specialized equipment, like 360-degree cameras, making it a highly professional endeavour. An alternative, yet underexplored, approach is outpainting existing movies into 360-degree formats.

Image-based panorama outpainting~\cite{tang2023MVDiffusion, wu2023panodiffusion, oh2022bips, akimoto2022diverse, chen2022text2light, zheng2022bridging, ying2020180, akimoto2022diverse, sumantri2020360} is a necessary step towards video-based 360 movie creation. The advance of text-to-image diffusion models~\cite{ramesh2022hierarchical, saharia2022photorealistic, rombach2022high, reed2016generative, ramesh2021zero, ho2020denoising} makes it possible to extrapolate an image into 360-degree view. PanoDiffusion~\cite{wu2023panodiffusion} proposes a panorama outpainting methods by fine-tuning a pretrained latent diffusion model~\cite{rombach2022high}. However, with a limited amount of training data available, this method disrupts the pre-trained model's prior knowledge and diminishing its generalization capabilities. MVDiffusion~\cite{tang2023MVDiffusion}, in contrast, maintains generalization by generating multi-view consistent panoramic images using a frozen pre-trained latent diffusion model. This method ensures geometric consistency through correspondence-aware attention, but it requires the input image to have known intrinsic and rotation matrices, limiting its application to panoramas from arbitrary images.

Extending panorama generation to camera-free inputs poses significant challenges. Our method estimates the input camera parameters by estimating the homography transformation from the input image to a predefined canonical view. The homography establishes a correspondence between input views and each panoramic views, allowing for the enforcement of multi-view consistency via correspondence-aware attention~\cite{tang2023MVDiffusion}. Furthermore, the homographic matrix estimation network is seamlessly integrated with the multi-view diffusion model in a fully differentiable manner. By doing so, the mechanism can effectively mitigate the errors introduced by the homography estimation process.

We fine-tuned our model on high-quality Matterport3D from the pre-trained stable diffusion inpainting model. We demonstrated the strong robustness of our model to random camera parameters. Our model also shows excellent generalization ability when tested on out-of-domain data from Structured3D without fine-tuning. 

To summarize, we make the following contributions:
\begin{itemize}
    \item We propose CamFreeDiff, a panorama generation model capable of handling camera-free input images. To the best of our knowledge, our model is the first of all designed to handle unkown input view and camera parameters.
    \item We formulate the camera parameter estimation by predicting the homography transformation from the input image to a predefined canonical view. We propose a three degrees of freedom(3-DoF) parameterization of homography instead of a standard 8-DoF way in the context of panorama outpainting. 
    \item We propose a novel strategy for incorporating camera prediction into the panorama generation model to enforce multi-view consistency and high visual quality.
\end{itemize}

%% file: text/2-related_work.tex
\section{Related Work}

\noindent \textbf{Multi-view image generation.}
The development of large-scale image generation models~\cite{saharia2022photorealistic, ramesh2022hierarchical, rombach2022high} has paved the way for multi-view image generation~\cite{shi2023mvdream, tang2023MVDiffusion, wang2023imagedream, liu2023syncdreamer, long2023wonder3d, tang2024mvdiffusion++, liu2023zero, shi2023zero123++}. A notable advancement in this area is the Zero-1-to-3 model~\cite{liu2023zero}, which can deduce novel views of an object given a single image and specified camera pose. Building on this, SyncDreamer~\cite{liu2023syncdreamer} introduces a method for generating images at fixed viewpoints through a combination of volumetric representation and epipolar attention, though it occasionally produces artifacts. In an effort to enhance consistency and image fidelity, models such as MVDream~\cite{shi2023mvdream}, ImageDream~\cite{wang2023imagedream}, and Wonder3D~\cite{long2023wonder3d} have employed comprehensive self-attention mechanisms across all views. Despite their success, these models necessitate extensive fine-tuning of a latent diffusion model's parameters, which can compromise pre-trained priors. While our method also adopts multi-view generation paradigms, our method is able to outpaint panorama for arbitrary images by freezing the pretrained stable diffusion model.   

\noindent \textbf{Panorama generation.}
Previous methods~\cite{chen2022text2light,stan2023ldm3d, wu2023panodiffusion, tang2023MVDiffusion, akimoto2019360, somanath2021hdr, zheng2019pluralistic} for generating panoramas typically rely on generative adversarial networks~\cite{goodfellow2020generative}, auto-regressive~\cite{van2016conditional} or diffusion models. Text2Light~\cite{chen2022text2light} emerged as a pioneering approach, utilizing a cascade auto-regressive model for panorama creation, whereas LDM3D~\cite{stan2023ldm3d} and PanoDiffusion~\cite{wu2023panodiffusion} have adopted latent diffusion techniques for generating or outpainting panoramas. Despite their innovations, these models often encounter a reduction in their ability to generalize when they are fine-tuned on limited datasets. On the other hand, the process of generating panoramic images inherently benefits from the direct correspondence between different views. Utilizing this, MVDiffusion~\cite{tang2023MVDiffusion} presents a novel approach by incorporating correspondence-aware attention to maintain multi-view consistency, without modifying the underlying pretrained model. 

However, challenges persist with current outpainting techniques, notably their struggle to generate coherent panoramas due to their inability to infer camera parameters. PanoDiff\cite{wang2023360} takes a step forward to estimate the longitude and latitude angles of a view, however still assumes camera roll angle and intrinsic. To address these shortcomings, we introduce a novel, fully differentiable pipeline designed to extrapolate from a single camera-free image to a complete 360-degree panorama. This method aims to bridge the gap between existing panoramic generation techniques and the need for accurate, consistent, and high-quality panoramic imagery, leveraging the strengths of direct view correspondence while overcoming the limitations of camera parameter estimation.

%% file: text/3-preliminary.tex
\begin{figure}[t!]
    \centering
    \includegraphics[width=1.0\textwidth]{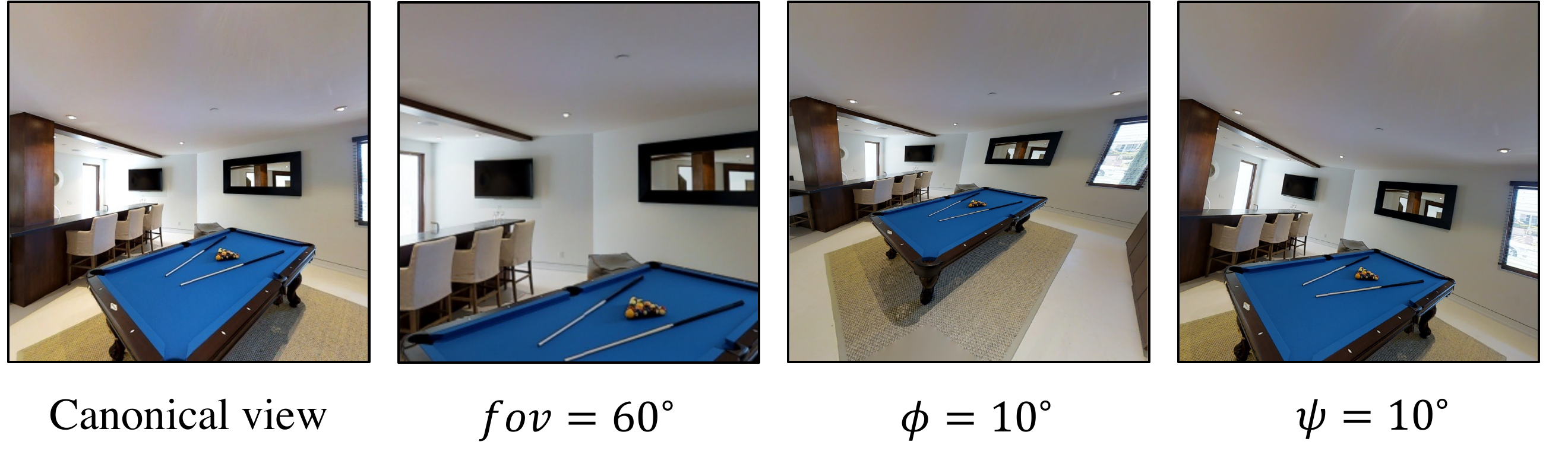}
    \caption{In the context of panorama generation, we define the camera parameter for a canonical perspective view as $fov=90^\circ$, $x$-axis rotation $\phi=0$ and $z$-axis rotation $\psi=0$. The $y$-axis rotation can be any $\theta$ from $0^\circ$ to $360^\circ$.}
    \label{fig:vis_view}
\end{figure}

\section{Preliminary}\label{sec:pre}
MVDiffusion is a multi-view diffusion model for coherent panoramic image generation. It splits the $360^\circ$ scene into eight perspective views $\textbf{I} = \{I_0, I_1, ..., I_7\}$ at fixed viewpoint, each with $90^\circ$ horizontal and vertical field of view and a $45^\circ$ horizontal overlapping. For image-based panorama outpainting, MVDiffusion employs a latent diffusion impainting model and makes the first view (canonical view), conditional on the input images. Correspondence-aware attention (CAA) blocks are designed to enforce geometry consistency. Consider a pair of corresponding points located at $p_s$ and $p_t$ in the source view $I_s$ and target view $I_t$ respectively, the information is aggregated from a $K \times K$ neighborhood $\mathcal{N}(p_s)$ centered at $p_s$ in the source view through a cross-attention mechanism:
\begin{equation}
M = \sum_{p^*_s \in \mathcal{N}(p_s)} \text{Softmax} \left(\ \left[ W_Q f(p_t) \right] \cdot \left[ W_K f(p^*_s) \right]\ \right)\ W_V f(p^*_s),
\end{equation}
Where $f(p_t)$ denotes image feature at location $p_t$ from target view and $f(p^*_s)$ represents image feature at location $p^*_s$ in the neighborhood $\mathcal{N}(p_s)$ from source view. $W_Q, W_K, W_V$ are the query, key and value projections for cross-attention in CAA blocks. MVDiffusion assumes the camera parameters of the input are the same as the canonical view, and thus cannot handle camera-free image.

%% file: text/4-method.tex
\begin{figure}[t!]
  \centering
  \includegraphics[height=6.5cm]{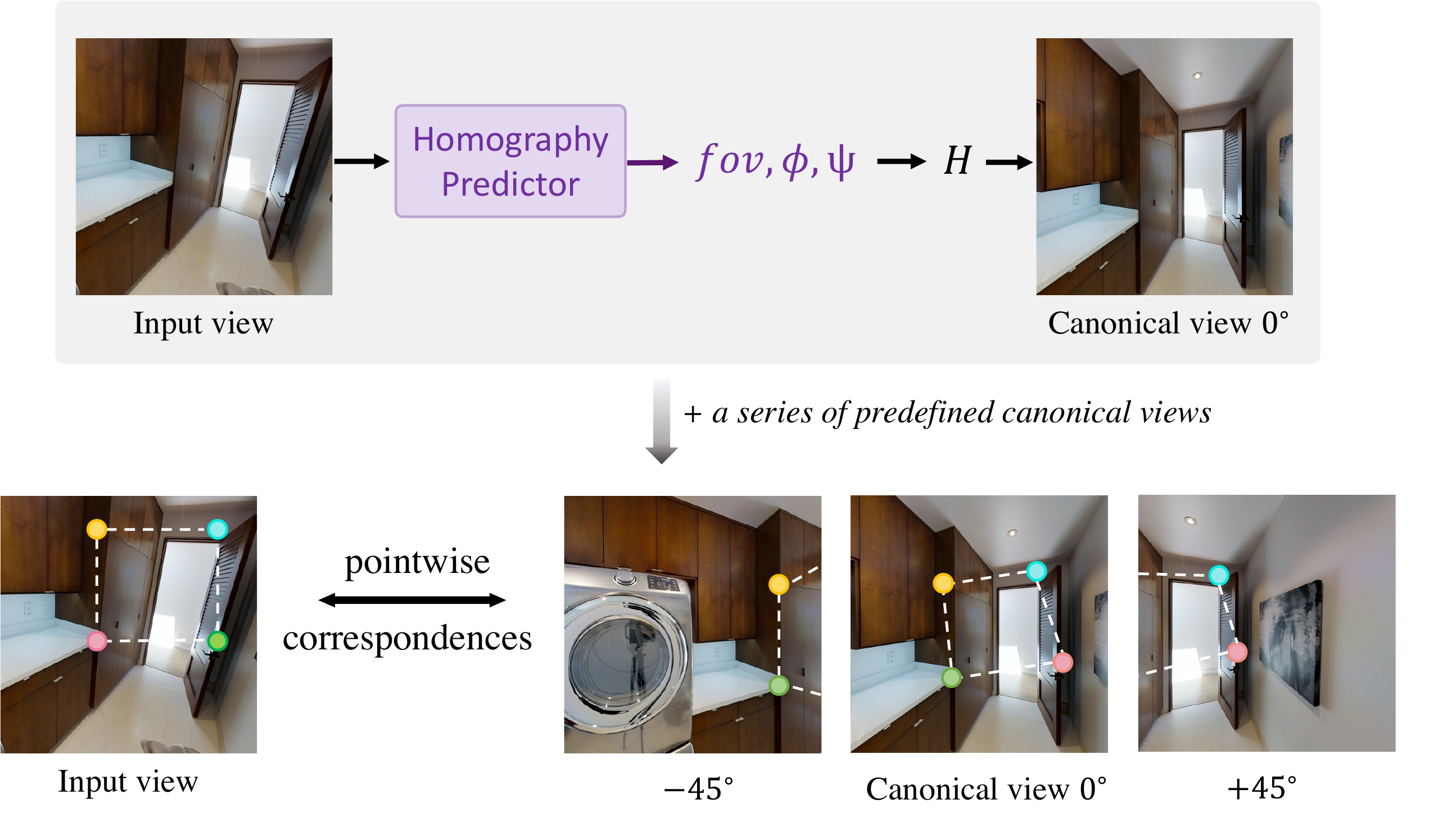}
  \caption{We formulate the camera parameter estimation as estimating the homography matrix from the input view to a predefined canonical view of the scene. We define the canonical view as the perspective view with an absolute rotation angle of 0$^{\circ}$.  We use a 3-DoF parameterization of the homography matrix instead of standard 8-DoF (details in \cref{sec:estimate_homography}).}
  \label{fig:homography}
\end{figure}

\section{Method}
We aim to generate a text-guided panoramic image from a camera-free input, where we assume the camera parameters are unknown. To address this problem, we propose our Camera-free Diffusion(CamFreeDiff) model to predict the camera parameters of the input image by estimating the homography transformation from the input to a predefined canonical perspective view. The predicted homography provides correspondences between multiple target perspective views and the input view, where correspondence-aware attention could be used to enforce geometry consistency for the final panorama generation.


\subsection{Homography Matrix Estimation}\label{sec:homo_est}
\label{sec:estimate_homography}
To generate a panorama based on the input image with an unknown camera, we first estimate the homography $H^{3 \times 3}$ matrix from the unknown input view to a predefined canonical view. \cref{fig:vis_view} Shows how we define a canonical view, and we illustrated our method in \cref{fig:homography}. 

\subsubsection{Parameterization of Homography Matrix.} Compared to the most straightforward way of parameterizing a homography into elements in the 3x3 matrix, DeTone \etal \cite{detone2016deep} demonstrated a 4-point parameterization of homography matrix can well reduce the optimization difficulty for a deep neural network, since the 4-points are independent of each other while the 3x3 matrix mixes both the rotational and the translational terms. 

In our case, parameters can be further simplified. When expressing homography as $H = K_2 R K_1^{-1}$, where $K_2$ and $K_1$ are the camera intrinsics for the canonical view and the input image accordingly, some variables are constant by default for common cameras. Specifically, the intrinsic matrix of a canonical view $K_2$ is
\begin{equation}
    \begin{bmatrix}
        f_x& \gamma &$\ $c_x \\
        0& f_y&$\ $c_y \\
        0& 0&$\ $1
    \end{bmatrix}
    =
    \begin{bmatrix}
        256& 0&$\ $ 255.5 \\
        0& 256&$\ $ 255.5 \\
        0& 0&$\ $1
    \end{bmatrix}
\end{equation}

We also assume the intrinsic of input image satisfies the pinhole camera and thus for $K_1$, in accord with our canonical view intrinsic, the axis skew coefficient $\gamma$ for input view also defaults to $0$ and principal point offsets ($c_x, c_y$) default to the centre of the image ($w/2,\ h/2$). 

Accordingly, for our task to generate 360-degree panorama, the homography matrix from input view to the canonical view has three degrees of freedom (3-DoF): 1) camera field of view ($f$), 2) camera rotation around the $x$-axis ($\phi$), 3) camera rotation around the $z$-axis ($\psi$). Particularly, under the condition of a single input image, predicting the absolute rotation around the $y$-axis ($\theta$) is considered meaningless since the input view can be mapped to any standard canonical view of a 360-degree panorama with $0 \leq \theta \leq 360^\circ$.

Therefore, we formulate the model that predicts homography matrix from an input image $I \in \mathbb{R}^{H\times W\times 3}$ to a predefined canonical view as:

\begin{equation}
\mathcal{M}(I) \rightarrow (\textit{f},\ \phi,\ \psi)
\end{equation}

The input image camera intrinsic $K_1$ can be determined by predicted $\textit{f}$. Along with known target perspective camera intrinsic and predicted rotation $R$ from ($\phi,\ \psi,\ \theta = 0$), the homography transformation $H$ can be recovered from the predictions $(f, \phi,\ \psi)$.

\subsubsection{Homography Estimator}
We build an MLP classifier with three hidden layers upon a general image encoder. The U-Net encoder pre-trained by stable diffusion model for image generation also serves as our homography estimation image encoder, but with weights frozen for efficiency. Only the MLP classifier is optimized to learn a homography estimator. Feature dimensions for each hidden layer in MLP are set to 5120, 2560 and 1280. $SiLU$\cite{elfwing2018sigmoid} is used as activation functions in MLP block. Cross-entropy loss is applied as learning objectives to $fov$, $\phi$ and $\psi$, respectively.

\begin{figure}[t!]
  \centering
    \includegraphics[width=1.0\linewidth]{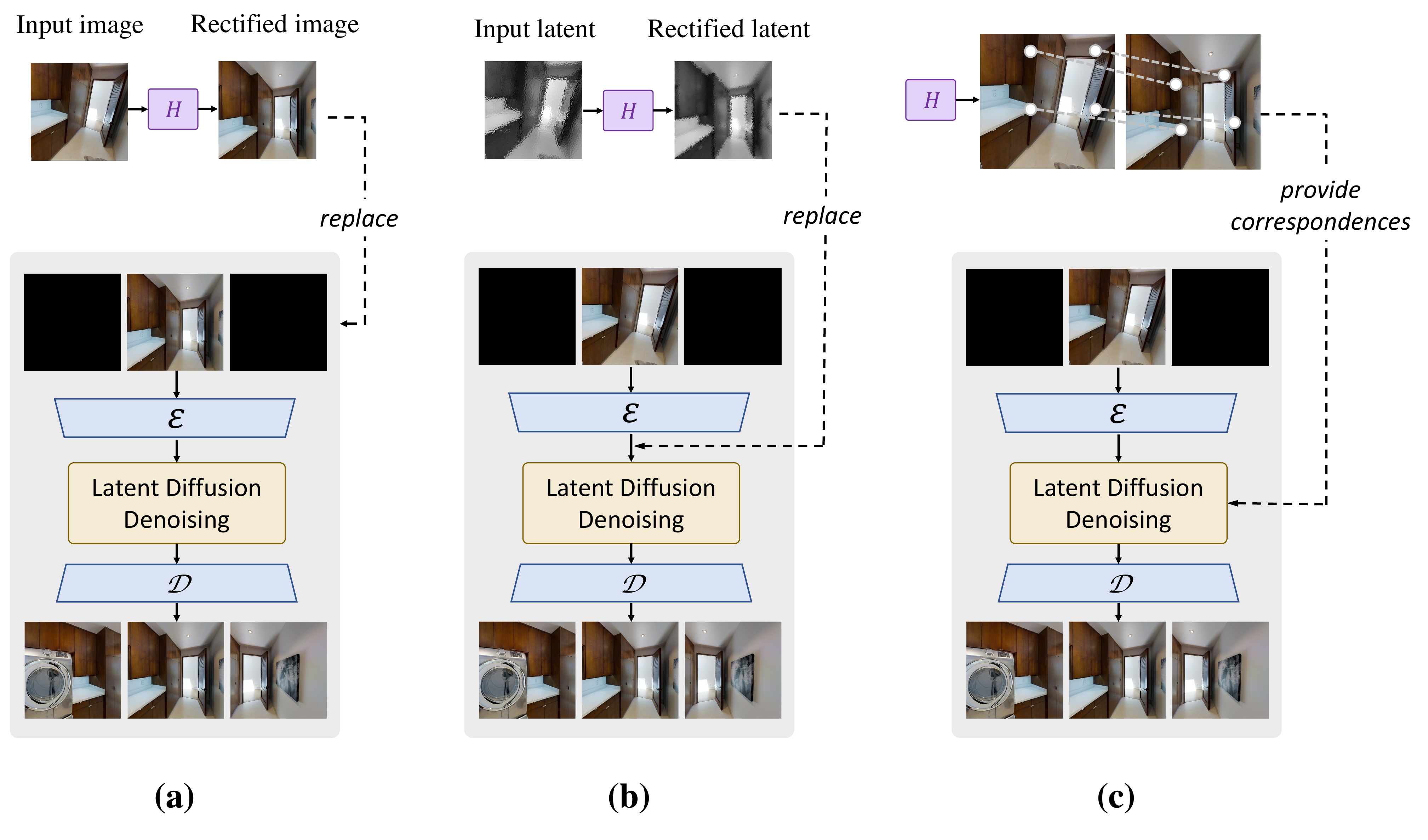}
    \caption{Different strategies to generate panorama from a camera-free input. After estimating the homography matrix $H$ from the input view to a predefined canonical view, alternatives are: (a) Rectify the input view by unwarping and replacing the original input image. (b) Rectify input view by unwarping and replacing the input latent after image encoding. (c) Provide point-level correspondences to the multi-view generation model to enforce consistency between corresponding points.}
  \label{fig:pipeline_warp}
\end{figure}

\subsection{Camera-free 360-Degree Image Outpainting}
\label{sec:method_outpaint}
Homography matrix from the camera-free input image to a predefined canonical image provides pixel-level correspondences between them. Consider the homography transformation from view $I_a$ to view $I_b$ as $H_{a,b}$. The projection from a point at location $p_a = (u_a,\ v_a)$ in view $I_a$ and its corresponding point at location $p_b = (u_b,\ v_b)$ in view $I_b$ is formulated as:

\begin{equation}
\begin{bmatrix}
u_b \\ v_b \\ 1
\end{bmatrix} = H_{a,b}\begin{bmatrix}
u_a \\ v_a \\ 1
\end{bmatrix} 
\end{equation}

Based on the estimated correspondences, we design three variants to generate 360-degree panoramic images, as shown in \cref{fig:pipeline_warp}.

\noindent \textbf{Variant 1(unwarp image):} Initially, the input image is transformed to a canonical view utilizing estimated correspondences, as depicted in Fig.~\ref{fig:pipeline_warp}(a). Subsequently, we incorporate the latent diffusion inpainting model to generate eight panoramic views (eight diffusion branches with the same weight copy), following the design of MVDiffusion~\cite{tang2023MVDiffusion} image-based panorama outpainting model. For the branch of canonical view $0^\circ$, input comprises a concatenation of noisy latent, the latent of the unwarped image, and a binary mask that identifies the areas requiring inpainting (\textit{zero} for the visible region and \textit{one} for the region to inpaint). The inputs for the remaining seven branches consist of the noisy latent, the latent of a purely white image, and a uniformly one-valued mask. It is important to note that the latent diffusion inpainting model is designed to preserve existing image content where the mask value is set to \textit{zero}, and to generate new content in areas where the mask value is \textit{one}.

\noindent \textbf{Variant 2(unwarp latent):} Unlike Variant 1, which initiates the process by unwarping the input image and subsequently encoding the unwarped image into a latent, in Variant 2, the image is first encoded into a latent space, followed by unwarping this latent into the canonical view $0^\circ$. Aside from this distinction, all other aspects of the design remain consistent with those outlined for Variant 1 as shown in \cref{fig:pipeline_warp}(b).

\begin{figure}[t!]
  \centering
  \includegraphics[width=0.9\linewidth]{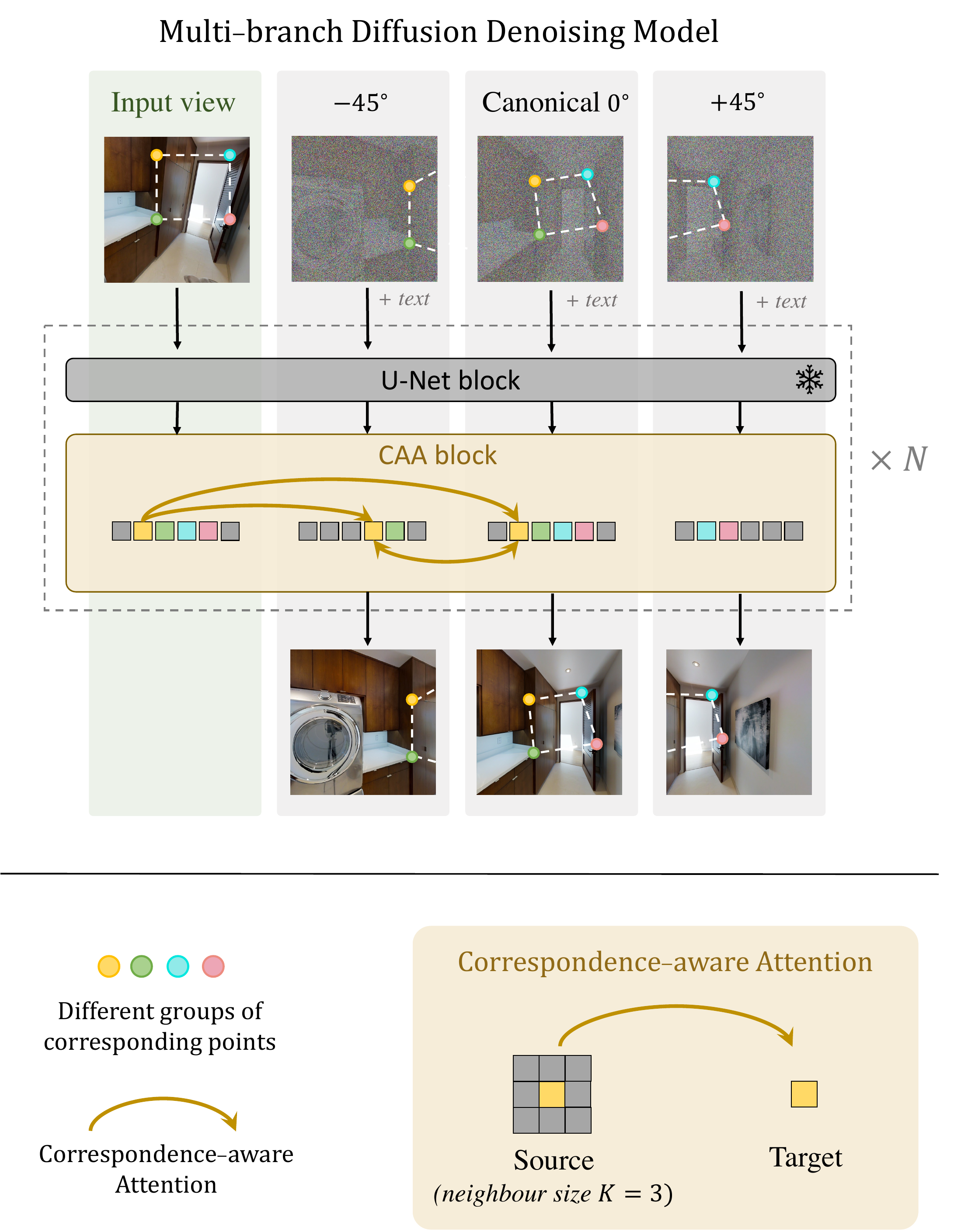}
  \caption{Our panorama generation pipeline based on multi-view diffusion denoising model. With the predicted homography matrix from the input view to a predefined canonical view, point-wise information can be aggregated from the input view to all target canonical views through correspondence-aware attention(CAA). Note that this figure only shows the cross-attention between one group of corresponding points for clear visualization. However, the same process is applied to all groups of corresponding points.}
  \label{fig:pipeline}
\end{figure}

\begin{figure}[t!]
  \centering
  \includegraphics[width=1.0\textwidth]{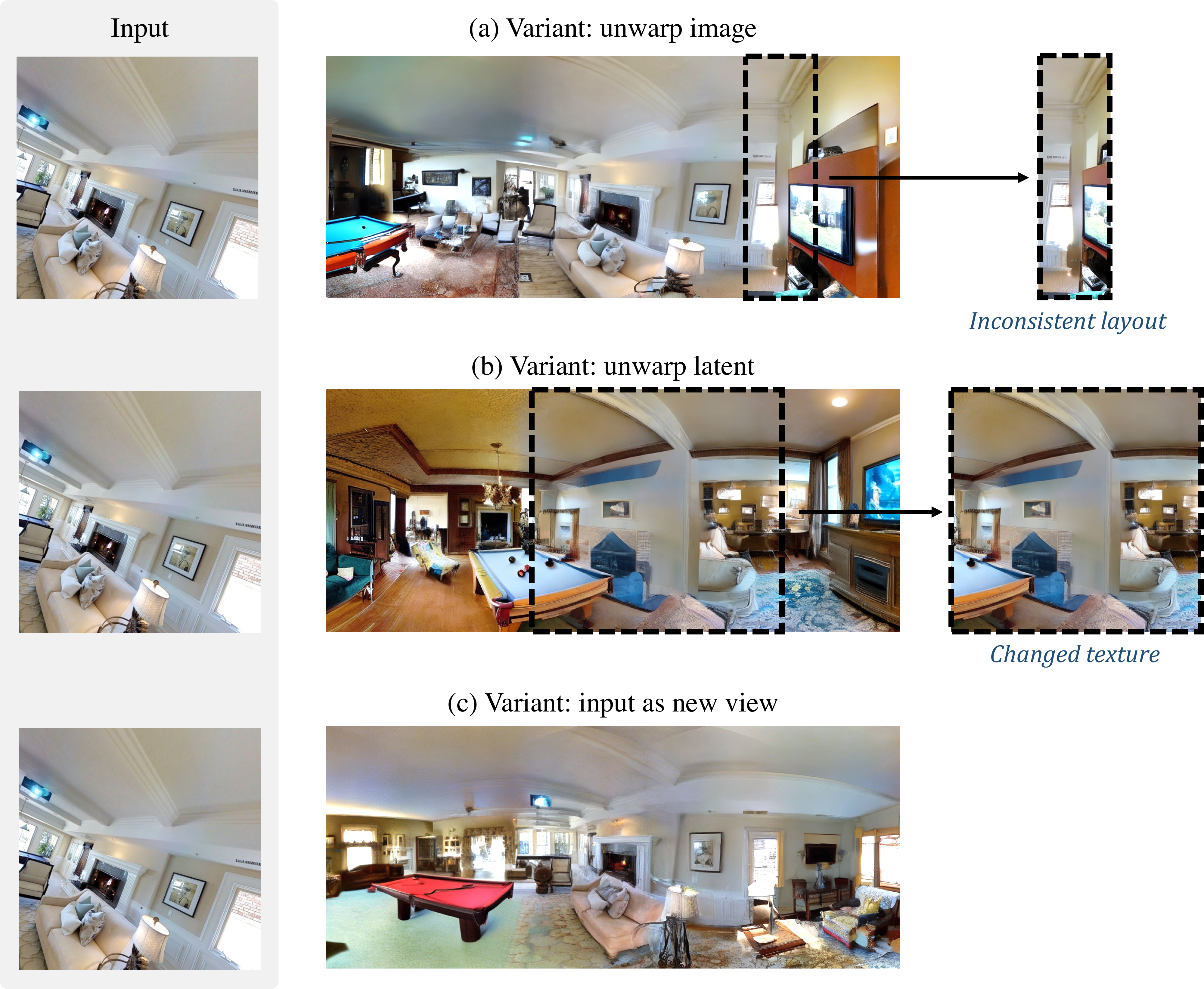}
  \caption{Qualitative comparison between different panorama generation strategies. For hard samples, as shown above, The variants involving unwarping image and unwarping latent (Variant 1 and 2) could lead to different issues.}
  \label{fig:variant_visual_compare}
\end{figure}

\begin{figure}[t!]
  \centering
  \includegraphics[width=1.0\textwidth]{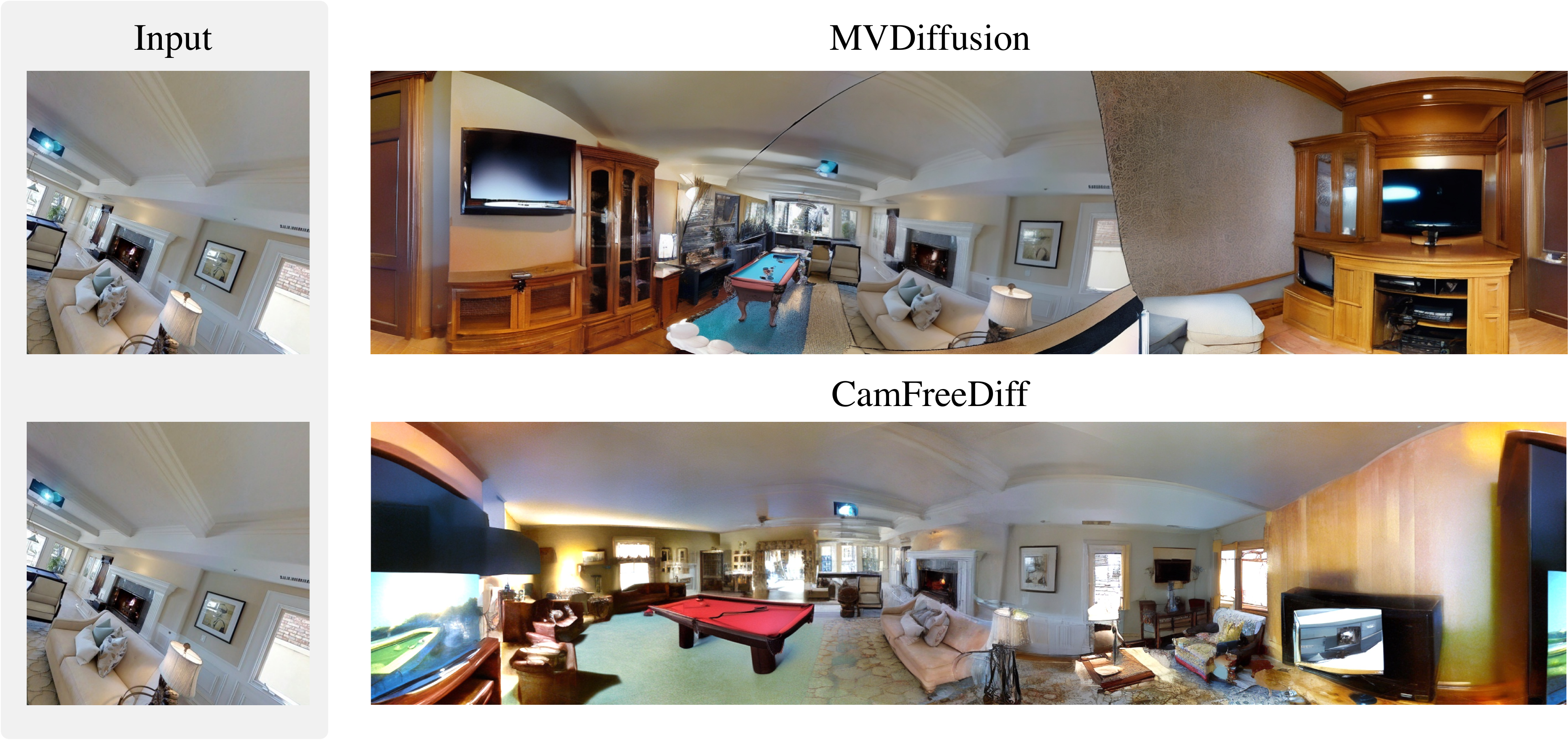}
  \caption{Comparison between baseline MVDiffusion and proposed CamFreeDiff. Note that we provide camera parameter estimation results to MVDiffusion since it has not been trained to handle arbitrary camera views. We observe that CamFreeDiff still shows stronger robustness to camera-free input.}
  \label{fig:compare_mvdiff}
  \vspace{2em}
\end{figure}

\noindent \textbf{Variant 3(new view):} Variant 3 introduces a novel approach that diverges from the strategies that involve unwarping images or latent employed in variants 1 and 2. This design is structured into one conditional branch and eight generation branches. Unlike Variants 1 and 2, where the canonical view $0^\circ$ branch depends on unwarped images or latent, the branch of canonical view is a generation branch in Variant 3, while the conditional branch depends on the input image. Input settings for other branches align with those specified in Variant 1. Particularly, Variant 3 enforces consistency between views by implementing CAA not only among the generation branches but also between the conditional branch and the generation branchs of canonical views. This strategy enables the effective reduction of the inaccuracies associated with homography estimation. We also illustrate our Variant 3 in \cref{fig:pipeline_warp}(c).






%% file: text/5-experiments.tex
\section{Experiments}
We provide experiment details in this section, including experiment setup (\ref{sec:exp_setup}, baselines (\ref{sec:baselines}), and experiment results (\ref{sec:results}).

\input{tables/table_main}
\input{tables/table_s3d}

\subsection{Experiment setup}\label{sec:exp_setup}
\noindent \textbf{Dataset.} We fine-tune our CamFreeView model on the real-world Matterport3D\cite{chang2017matterport3d} dataset, which contains 90 building-scale indoor scenes with 10,912 high-resolution panoramic images. 9820 and 1092 images are split for training and evaluation, respectively, following MVDiffusion~\cite{tang2023MVDiffusion}. Each room in the dataset provides six distinct non-overlapping perspective images taken from identical camera positions, with each offering a $90^\circ$ field of view. To reach our goal of learning 360-degree image-to-panorama outpainting from camera-free input, we apply a random warp on each perspective image with a field of view from $60^\circ$ to $110^\circ$ and camera rotations of $\pm 15^\circ$ to create camera-free images. After random warping, all input images are in $512 \times 512$ resolution. 

In addition to the primary dataset, we conducted further evaluation of our model on the Structured3D\cite{zheng2020structured3d} dataset, a photo-realistic compilation of 3,500 indoor scenes encompassing 21,835 rooms, each rendered with panoramic images. Perspective images for each room are also generated from random camera positions and poses. This step aims to rigorously assess the model's generalization abilities on out-of-domain data. We applied the same random warp approach in Structured3D as in Matterport3D.

BLIP-2 \cite{li2023blip} captioning model is used to generate per-view text descriptions for both datasets mentioned above.

\noindent \textbf{Training Details.} Our CamFreeDiff model is fine-tuned from the stable diffusion inpainting model. We retain the weights of VAE image encoder/decoder and the latent denoising U-Net blocks frozen as pre-trained. We optimize the MLP block for homography prediction and the CAA blocks for multi-view consistency with a learning rate $2\times 10^{-4}$ for 30 epochs. 

\noindent \textbf{Evaluation Metrix.} Our text-guided 360-degree image outpainting employs a series of standard image generation metrics to evaluate visual quality: Frechet Inception Distance (FID)\cite{heusel2017gans}, which quantifies the distance between real and generated images; Inception Score (IS)\cite{salimans2016improved} offering insight into the diversity and quality of generated images; and CLIP score\cite{radford2021learning} to measure the alignment between text descripition and corresponding images. In addition, we use the Peak signal-to-noise ratio (PSNR) on the corresponding region between the generated and target canonical view $0^\circ$ to evaluate our view estimation error. We also use Mean Absolute Error to assess the accuracy of homography estimation only.

\begin{figure}[t!]
  \centering
    \includegraphics[width=0.9\linewidth]{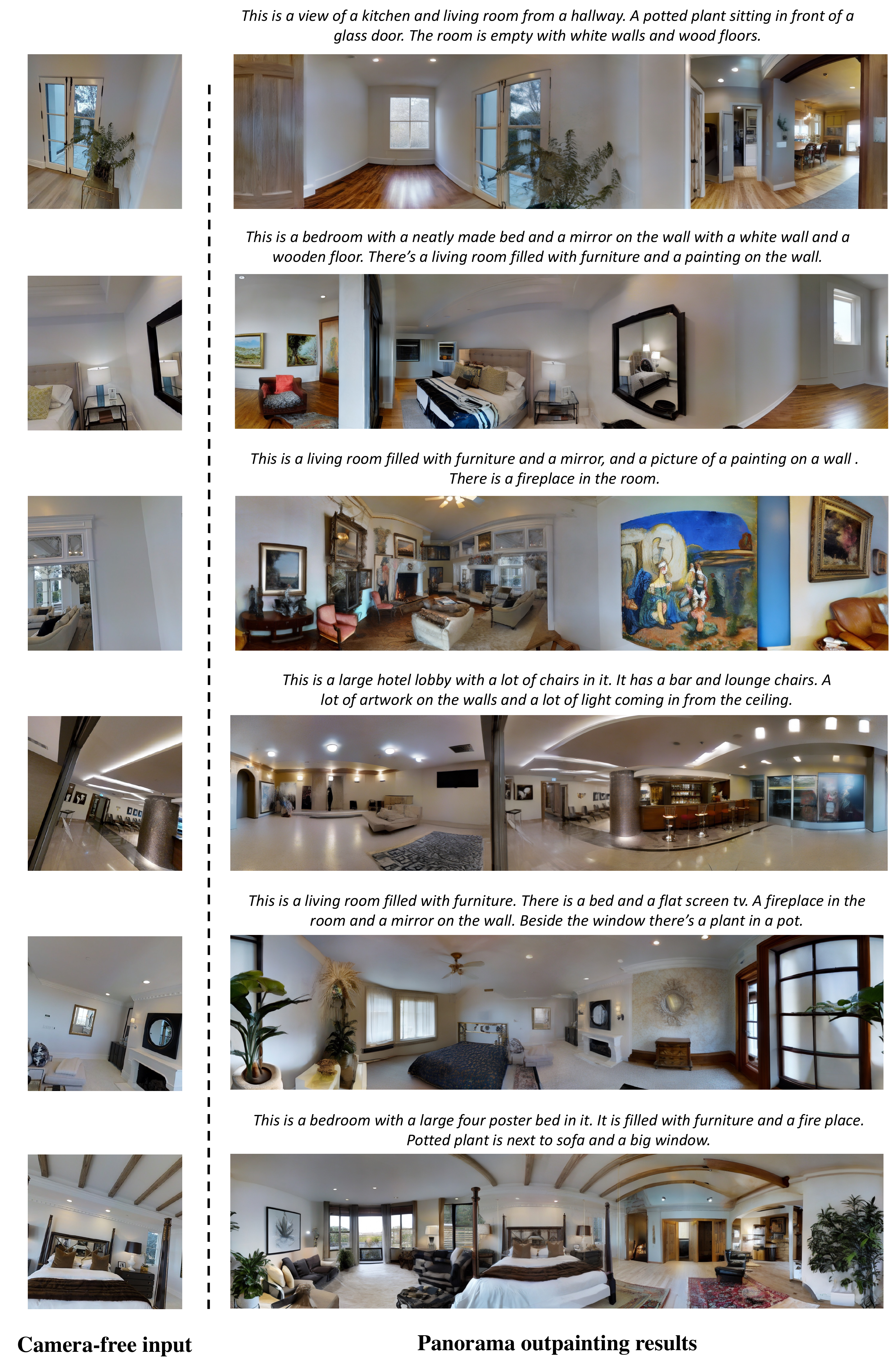}
    \caption{More qualitative results of camera-free panorama generation from our CamFreeDiff model.}
  \label{fig:vis_results}
\end{figure}

\subsection{Baselines} \label{sec:baselines}
We consider the following baselines: MVDiffusion and PanoDiffusion. 

\textit{MVDiffusion} is a multi-view text-to-image diffusion model to generate view-consistent 360-degree scenes. For comparison, our model learns to generate from camera-free input with unknown camera parameters. 

\textit{PanoDiffusion} is designed for RGB-D panorama outpainting with different types of masks. A super-resolution model further enhances the outpainting results with higher resolution. 

To the best of our knowledge, no existing panorama generation model was designed or learned to handle input images from different camera poses (rotation) and focal lengths (field of view). Therefore, by default, we provide all baselines with rectified input images using predicted camera parameters by our homography estimator for fair comparison. In addition, we also compare our three variants.

\input{tables/ablate_homoNet}


\subsection{Results} \label{sec:results}
\subsubsection{Qualitative Results}
In \cref{fig:compare_mvdiff} we show qualitative comparison between baseline MVDiffusion and CamFreeDiff. Both MVDiffusion and CamFreeDiff are trained on Matterport3D dataset. CamFreeDiff is designed and trained to handle arbitrary camera parameters, while MVDiffusion is not.  
We report qualitative results for different variants of outpainting strategies as introduced in \cref{{sec:method_outpaint}}. Qualitative comparison for a hard sample is shown in \cref{fig:variant_visual_compare}. We observe that the homography prediction error can be propagated into later panorama generation for Variant 1(unwarping image), resulting in inconsistent scene layout. For Variant 2(unwarping latent), inaccurate homography prediction could lead to changed textures for the input view. Compared to the above variants, introducing a new conditional branch for the input view and leveraging correspondence-aware attention(Variant 3) enables the most consistent and robust multi-view generation. We provide more qualitative results for CamFreeDiff in \cref{fig:vis_results}.

\subsubsection{Quantitative Results}
We show the numerical results for panorama generation between baselines and different variants of our CamFreeDiff model in \cref{tab:main}. Among them, CamFreeDiff with Variant 3, treating the input as a new view, achieved the best results in terms of visual quality for panorama generation (FID, IS, CS) and reconstruction quality (PSNR).

To demonstrate the generalization ability of our model, we also tested our model on the Structured3D\cite{zheng2020structured3d} dataset. Note that our model was never trained on or applied with domain transfer techniques from Structured3D. Results shown in \cref{tab:s3d} indicate the strong generalization ability of CamFreeDiff to out-of-domain data, even surpassing PanoDiffusion, which is trained directly on Structured3D but without learning from camera-free input.

\input{tables/ablate_homo_obj}

\subsection{Ablation Study}
\subsubsection{Homography Estimator} We compared classification and regression as different types of homography estimators. Cross-entropy loss is used as the objective for the classifier, and mean squared error (MSE) loss is for the regression model. The classifier gives the best input view estimation results instead of the regression model, as shown in table \cref{tab:ablate_homo_obj}. In addition, we ablate on the architecture design of the homography estimator. We compared our design of an MLP block built on a frozen stable diffusion image encoder with the HomographyNet, which is designed to predict homography matrix between views. Our designs achieve better generation results as shown in \cref{tab:ablate_homo_net}.

\input{tables/ablate_nsize}

\subsubsection{Neighborhood Size in CAA blocks} Given correspondences between views, Correspondence-aware attention (CAA) aggregates information from source point $p_s$ neighborhood to target point $p_t$, which is the key to yielding consistency between multiple views. The neighborhood in CAA refers to the $K \times K$ neighboring points centered at $p_s$. We ablate the neighborhood size $K$ for $K = 1,3,5,7$. From results shown in \cref{tab:ablate_neighbor_size}, we find larger neighborhood size generally leads to better multi-view generation quality, but the improvement is limited. Note that larger $K$ results in more computational and time complexity for CAA operation.

%% file: tables/table_main.tex
\begin{table}[t!]
  \caption{We report the panorama generation results of our proposed CamFreeDiff on Matterport3D dataset. For baseline PanoDiffusion and MVDiffusion, we use our homography predictor to rectify the camera-free input image first, and then evaluate their panorama generations only. Therefore, PSNR is not reported for these baselines. }
  \label{tab:main}
  \centering
  \begin{tabular}{p{0.25\textwidth}
  >{\centering\arraybackslash}p{0.1\textwidth}
  >{\centering\arraybackslash}p{0.1\textwidth}
  >{\centering\arraybackslash}p{0.1\textwidth}
  >{\centering\arraybackslash}p{0.1\textwidth}}
    \toprule
    Model & FID $\downarrow$ & IS $\uparrow$ & CS$\uparrow$ & PSNR$\uparrow$ \\
    \midrule
    PanoDiffusion & 48.7 & 3.1 & - & - \\
    MVDiffusion & 42.4 & 5.4 & 21.9 & - \\
    \midrule
    \textbf{CamFreeDiff} &  &  & \\
    + unwarp image & 35.2 & 5.5 & 23.6 & 18.7 \\
    + unwarp latent & 34.3 & 5.6 & 22.4 & 15.6 \\
    \textbf{+ new view} & \textbf{27.0} & \textbf{5.6} & \textbf{24.4} & \textbf{19.3} \\
  \bottomrule
  \end{tabular}
\end{table}

%% file: tables/table_s3d.tex
\begin{table}[t!]
  \caption{We test our model on the Structured3D(S3D) dataset without applying fine-tuning or domain transfer techniques to demonstrate the generalization ability. Note that PanoDiffusion is trained on Structured3D without text guidance. Therefore, we provide no CLIP score (CS).}
  \label{tab:s3d}
  \centering
  \begin{tabular}{>{\centering\arraybackslash}p{0.25\textwidth} |
  >{\centering\arraybackslash}p{0.12\textwidth} 
  >{\centering\arraybackslash}p{0.12\textwidth} |
  >{\centering\arraybackslash}p{0.08\textwidth}
  >{\centering\arraybackslash}p{0.08\textwidth}
  >{\centering\arraybackslash}p{0.08\textwidth}
  >{\centering\arraybackslash}p{0.08\textwidth}}
    \toprule
    Model & Training & Cam-free & FID $\downarrow$ & IS $\uparrow$ & CS$\uparrow$ & PSNR$\uparrow$ \\
    \midrule
    PanoDiffusion & \twemoji{check mark} & \twemoji{multiply} & 35.3 & 3.2 & - & - \\
    \midrule
    \textbf{CamFreeDiff} & \twemoji{multiply} & \twemoji{check mark} & \textbf{31.1} &\textbf{ 6.2} & 25.5 & 19.8  \\

  \bottomrule
  \end{tabular}
\end{table}

%% file: tables/ablate_homoNet.tex
\begin{table}[t!]
  \caption{Comparison of the architecture design for homography estimator.
  }
  \label{tab:ablate_homo_net}
  \centering
  \begin{tabular}{p{0.35\textwidth}
  >{\centering\arraybackslash}
  >{\centering\arraybackslash}p{0.08\textwidth}
  >{\centering\arraybackslash}p{0.08\textwidth}
  >{\centering\arraybackslash}p{0.08\textwidth}
  >{\centering\arraybackslash}p{0.12\textwidth}}
    \toprule
    Model & FID $\downarrow$ & IS $\uparrow$ & CS$\uparrow$ & PSNR$\uparrow$ \\ 
    \midrule
    HomographyNet & 29.2 & 5.6 & 24.2 & 19.2 \\
    SD encoder(frozen) + MLP & \textbf{27.0} & 5.6 & \textbf{24.4} & \textbf{19.3} \\
    %
  \bottomrule
  \end{tabular}
\end{table}

%% file: tables/ablate_homo_obj.tex
\begin{table}[t!]
  \caption{Ablation study of different types of homography estimation objectives. Experiments are based on HomographyNet and evaluated on randomly warped images from Matterport3D. Predictions are evaluated in terms of Mean Absolute Error.}
  \label{tab:ablate_homo_obj}
  \centering
  \begin{tabular}{>{\centering\arraybackslash}p{0.2\textwidth}
   |
  >{\centering\arraybackslash}p{0.1\textwidth}
  >{\centering\arraybackslash}p{0.1\textwidth}
  >{\centering\arraybackslash}p{0.1\textwidth}}
    \toprule
    Objectives & fov$ \downarrow$ & phi($\phi$)$ \downarrow$ & psi($\psi$)$ \downarrow$ \\
    \midrule
    MSE (reg.)  & 10.6 & 2.5 & 2.4 \\
    CE (cls.) & \textbf{7.9} & \textbf{1.8} & \textbf{1.5} \\
  \bottomrule
  \end{tabular}
\end{table}



%% file: tables/ablate_nsize.tex
\begin{table}[t!]
  \caption{Ablation study on the neighbour size $K$ in CAA blocks for information aggregation between two views. 
  }
  \label{tab:ablate_neighbor_size}
  \centering
  \begin{tabular}{>{\centering\arraybackslash}p{0.2\textwidth} | >{\centering\arraybackslash}p{0.08\textwidth} >{\centering\arraybackslash}p{0.08\textwidth} >{\centering\arraybackslash}p{0.08\textwidth} >{\centering\arraybackslash}p{0.08\textwidth}}
    \toprule
    Neighbour Size & FID $\downarrow$ & IS $\uparrow$ & CS$\uparrow$ & PSNR$\uparrow$ \\
    \midrule
    $K=1$ & 66.4 & 5.5 & 22.3 & 19.2 \\
    $K=3$ & 51.1 & 5.8 & 23.9 & \textbf{19.5} \\
    $K=5$ & 48.0 & \textbf{6.2} & \textbf{24.1} & 19.4 \\
    $K=7$ & \textbf{47.3} & 6.1 & \textbf{24.1} & 19.4 \\
  \bottomrule
  \end{tabular}
\end{table}

%% file: text/6-conclusion.tex
\section{Conclusion}

This paper introduces a novel method for generating panorama from a camera-free input image. We formulate the camera parameter estimation of the input as an estimation of the homography matrix from the input view to a predefined canonical view of the scene. Our method builds upon the MVDiffusion model for multi-view image generation and incorporates correspondences between the input and target canonical views for coherent and consistent panorama generation. We demonstrate the strong robustness of our model to camera-free inputs and generalization ability to out-of-domain data.